\documentclass[conference]{IEEEtran}
\IEEEoverridecommandlockouts
\usepackage{cite}
\usepackage{amsmath,amssymb,amsfonts}
\usepackage{algorithmic}
\usepackage{algorithm}
\usepackage{graphicx}
\usepackage{textcomp}
\usepackage{xcolor}
\usepackage{makecell}
\usepackage{multirow}
\usepackage{multicol}
\usepackage{float}
\usepackage{enumitem}
\usepackage{caption}
\usepackage{subcaption}
\usepackage{hyperref}
\hypersetup{
colorlinks,
citecolor=blue,
linkcolor=blue
}

\def\BibTeX{{\rm B\kern-.05em{\sc i\kern-.025em b}\kern-.08em
    T\kern-.1667em\lower.7ex\hbox{E}\kern-.125emX}}
\begin{document}

\title{Power grid operational risk assessment using graph neural network surrogates \\

\thanks{This study was partly funded by the Advanced Research Projects Agency-Energy (ARPA-E) Perform Award AR0001136.}
}

\author{\IEEEauthorblockN{Yadong Zhang}
\IEEEauthorblockA{\textit{Dept. of Civil and Environmental {Engg.}} \\
\textit{Vanderbilt University}\\
Nashville, United States \\
yadong.zhang@vandebilt.edu}
\and
\IEEEauthorblockN{Pranav M Karve}
\IEEEauthorblockA{\textit{Dept. of Civil and Environmental {Engg.}} \\
\textit{Vanderbilt University}\\
Nashville, United States \\
pranav.m.karve@vanderbilt.edu}
\and
\IEEEauthorblockN{Sankaran Mahadevan}
\IEEEauthorblockA{\textit{Dept. of Civil and Environmental {Engg.}} \\
\textit{Vanderbilt University}\\
Nashville, United States \\
sankaran.mahadevan@vanderbilt.edu}
}

\maketitle

\begin{abstract}
We investigate the utility of graph neural networks (GNNs) as proxies of power grid operational decision-making algorithms (optimal power flow (OPF) and security-constrained unit commitment (SCUC)) to enable rigorous quantification of the operational risk. To conduct principled risk analysis, numerous Monte Carlo (MC) samples are drawn from the (foretasted) probability distributions of spatio-temporally correlated stochastic grid variables. The corresponding OPF and SCUC solutions, which are needed to quantify the risk, are generated using traditional OPF and SCUC solvers to generate data for training GNN model(s). The GNN model performance is evaluated in terms of the accuracy of predicting quantities of interests (QoIs) derived from the decision variables in OPF and SCUC. Specifically, we focus on thermal power generation and load shedding at system and individual zone level. We also perform reliability and risk quantification based on GNN predictions and compare with that obtained from OPF/SCUC solutions. Our results demonstrate that GNNs are capable of providing fast and accurate prediction of QoIs and thus can be good surrogate models for OPF and SCUC. The excellent accuracy of GNN-based reliability and risk assessment further suggests that GNN surrogate has the potential to be applied in real-time and hours-ahead risk quantification. 
\end{abstract}


\section{Background}

The growing participation of renewable sources (RES) together with flexible loads bring considerable volatility and uncertainty to grid operation~\cite{ghorani2019risk}. System operators have to employ advanced uncertainty quantification and propagation (UQ and UP) tools for risk-aware power grid operation, which includes power generation scheduling (day-ahead, hours-ahead) and dispatch. Optimal operational decision-making algorithms, such as optimal power flow (OPF) and security-constrained unit commitment (SCUC), play a key role in performing these tasks~\cite{zheng2019hierarchical,ramesh2021accelerated}. The OPF problem aims to minimize the total power generation cost while adhering to supply-demand equilibrium, reserve requirements, generation capacity limits, and network constraints, at a given time. It tries to find the most cost-effective power dispatch from available (online) generators to meet the immediate or \textit{real-time} load demand. SCUC, on the other hand, is a hours-ahead process that extends over a planning horizon to determine the optimal online/offline schedules of generators. It aims to bring generators online to fulfill the future demand at the lowest possible cost, incorporating constraints such as minimum up and down hours, ramping capabilities, etc.~\cite{kourounis2018toward}. The SCUC algorithm \emph{decides} which units are available for dispatch by determining their commitment status ahead of time, after which OPF takes over to manage the real-time dispatch of these units~\cite{paturet2020stochastic}. 

To evaluate the risk underlying a given unit commitment (UC) and dispatch decision, sampling-based approach has been proposed~\cite{stover2023reliability,zhang2023graph}. In this approach, numerous OPF and SCUC solutions (future decisions) corresponding to various possible future scenarios are obtained. The quantities of interest (QoIs) corresponding to these future scenarios and decisions are analyzed to quantify the risk. The need to solve numerous OPF/SCUC problems poses a significant computational challenge, since solution of these mixed-integer programming problems for even a single future scenario is computationally challenging. The advancement of machine learning (ML) has provided new opportunities to address this challenge. Among various ML model architectures, graph neural network (GNN) is particularly attractive for building a surrogate model for computational tasks in power grid, since it is capable of handling graph-structured data~\cite{wu2020comprehensive}. The fundamental idea of GNN is to learn the nodal features in a graph by assimilating information from its neighborhood nodes by leveraging message passing mechanism. GNN has been proved useful in diverse power grid operational tasks~\cite{liao2021review}, including fault diagnosis, power outage prediction, line flow control, distributed energy sources operation, transient stability assessment, etc. To the best of our knowledge, however, the utilization GNN in real-time and hours-ahead reliability and risk assessment with the inclusion of spatio-temporal dependency has not been previously reported. 

In this work, we investigate the possibility of using GNN as surrogate model for fast OPF and SCUC computation to enable the real-time and hours-ahead reliability and risk assessment. Note that the grid variables are not independent but manifest significant correlations. Fluctuation of variables in one zone is often related to the changes in other zones, leading to spatial correlation. Temporal correlations also emerge due to grid inertia and sequential nature of grid operations. These correlations must be considered in forecasting, OPF and SCUC computations, as well as in risk quantification. In this work, to model the spatial dependency of stochastic variables, (uncorrelated) Monte Carlo (MC) samples are drawn from the respective marginal distributions, and are then converted into correlated samples with specified correlation coefficient(s). Temporal dependency of stochastic variables is also considered by including autocorrelation function in generating MC samples (scenarios). These samples are then used to obtain numerical OPF and SCUC solutions. Following supervised learning approach, the GNN surrogate models are trained with scenarios of stochastic variables as input and corresponding numerical solutions as the ground truth. Model performance are evaluated with focus on the prediction accuracy of thermal power generation and load shedding, i.e. quantities of interests (QoIs). The predictions are then used to in reliability and risk quantification.


\section{Proposed Solution}

\subsection{GNN surrogate development}

The fundamental idea of GNN is to update nodal features in a graph by assimilating information from their neighboring nodes, i.e., message passing. Within the topology of a graph, nodes directly linked to a given node are designated as its \emph{one-hop} neighbors, and the immediate neighbors of these one-hop nodes are identified as \textit{two-hop} neighbors, and so on. Each layer of a GNN model facilitates exchange of information between a node and its one-hop neighbors, and multiple layers can be stacked together to assimilate information from farther neighbours. Message passing in the $k$-th layer can be stated as:
\begin{align}
    \mathbf{m}_u &= \text{\textbf{AGGREGATE}}(\mathbf{h}^k_v), \; v \in \mathcal{N}(u) \\
    \mathbf{h}_u^{(k+1)} &= \text{\textbf{UPDATE}}(\mathbf{h}_u, \; \mathbf{m}_u),
\end{align}
where $\mathbf{h}_u^k$ represents the embedding of node $u$ in $k$-th layer.

Various GNN architectures have been developed by changing aggregation and update functions, among which graph convolutional network (GCN) is one of the earliest architectures and has achieved remarkable success in many applications\cite{kipf2016semi}. GCN considers symmetric-normalized aggregation with self-loop added as:
\begin{align}
    \mathbf{m}_{u} = \sum_{v \in \mathcal{N}(u)\cup u}\frac{\mathbf{h}_v}{\sqrt{|\mathcal{N}(u)||\mathcal{N}(v)|}},
\end{align}
where $\mathcal{N}(u)$ is the set of neighbouring nodes of $u$, $| \cdot |$ represents the cardinality of a set. The update function can be set as mean, max, min, etc.

The GNN surrogates used in this article are constructed using three GCN layers followed by a single, fully connected readout layer. GCN layers are used to update node representation while the readout layer is used to output the QoI with the correct dimension. The loss function is constructed using the mean squared error (MSE) loss with a penalty term:
\begin{align}
    \label{MSE_loss}
    L = \frac{1}{|\mathcal{V}|}\sum_{i \in \mathcal{V}}(x_i-x_i^\ast)^2 + \frac{1}{|\mathcal{V}|}\sum_{i \in \mathcal{V}}\zeta_i^2,
\end{align}
where $\mathcal{V}$ denotes the set of nodes, $x_i^\ast$ and $x_i$ represent the ground truth an dprediction, respectively. $\zeta_i$ is an auxiliary variable defined as:
\begin{align}
    \label{penalty_term}
    \zeta_i = \text{max}\{x_i-x^{\rm max}_i, 0\} - \text{min}\{x-x^{\rm min}_i, 0\}
\end{align}
with $x^{\rm min}$ and $x^{\rm max}$ denoting the minimum and maximum allowed value of QoIs.

\subsection{Training data generation}

Grid variables manifest temporal correlation as well as randomness at individual time step. Here, a random walk is utilized to model the temporal dependency. Given state at $t$, the state at $(t+1)$ is represented as:
\begin{align}
    \mathbf{x}_{t+1} = \mathbf{x}_t + \mathbf{\epsilon}_{\mathcal{N}}, \;\; \mathbf{\epsilon}_{\mathcal{N}} \in \mathcal{N}(0, 1) \label{eq:random_walk}
\end{align}
Given the initial state $\mathbf{x}_1$, the subsequent states within a period are obtained using Eq.~\ref{eq:random_walk}. If $\mathbf{x}_1$ is described by standard normal, then subsequent state at $t$ is described by $\mathcal{N}(0, t)$. Herein, we convert $\mathbf{x}_t$ to standard normal using $ \mathbf{s}_t = \mathbf{x}_t / \sqrt{t}$. The (spatially) uncorrelated grid variables $\mathbf{s}_t$ are then transformed to correlated variables using:
\begin{align}
    \mathbf{x}^c_t = \left( \mathbf{L}\mathbf{s}_t^T \right)^T,
\end{align}
where $\mathbf{L}$ is the lower triangular matrix in Cholesky factorization of (forecast) covariance matrix $\mathbf{C}$. Given the marginal probability distributions $\mathcal{W}_i$, $i \in \mathbb{N}^+$ and covariance matrix $\mathbf{C}$, the spatially correlated variables from these marginals can be obtained as:
\begin{align}
    \mathbf{w}_t^i = \Phi^{-1}_{\mathcal{W}_i}\left( \mathbf{u}_t^i \right),
\end{align}
where $\mathbf{u}_t^i$ is the $i$-th column of $\mathbf{u}_t$, which is obtained from the CDF of standard normal $\mathbf{u}_t = \Phi\left(\mathbf{x}_t^c\right)$. $\Phi_{\mathcal{W}_i}$ represent the CDF of $i$-th marginal. The procedure used for generating spatio-temporally correlated stochastic grid variables from different marginals is summarized in Algorithm~\ref{alg:alg1}.

\begin{algorithm}[!ht]
\caption{Sampling spatio-temporally correlated stochastic grid variables}\label{alg:alg1}
\begin{algorithmic}

\STATE \textbf{Inputs}: $\#$samples $N$, $\#$time steps $T$, marginals $\mathcal{W}_1$, $\mathcal{W}_2$, ..., $\mathcal{W}_M$, covariance matrix $\mathbf{C} \in \mathbb{R}^{M \times M}$

\STATE \textbf{Start} 

\STATE \textbf{do}: $\mathbf{L} \gets  \mathbf{C} = \mathbf{L}\mathbf{L}^T$, $\mathbf{x}_0 \gets \mathbf{0} \in \mathbb{R}^{N \times M}$

\STATE \textbf{for} t = 1:T \textbf{do}:

\STATE \hspace{0.5cm} $\mathbf{x}_t \gets \mathbf{x}_{t-1} + \epsilon_{\mathcal{N}}$, $\epsilon_{\mathcal{N}} \sim \mathcal{N}(0, 1)$, $\epsilon_{\mathcal{N}} \in \mathbb{R}^{N \times M}$

\STATE \hspace{0.5cm} $\mathbf{s}_t \gets \mathbf{x}_t / \sqrt{t}$

\STATE \hspace{0.5cm} $\mathbf{x}^c_t \gets \left( \mathbf{L} \mathbf{s}_t^T \right)^T$

\STATE \hspace{0.5cm} $\mathbf{u}_t \gets \Phi\left(\mathbf{x}^c_t\right)$

\STATE \textbf{End for}

\STATE \textbf{for} t = 1:$T$ \textbf{do}:

\STATE \hspace{0.5cm} \textbf{for} i = 1:M \textbf{do}:

\STATE \hspace{1cm} $\mathbf{w}^i_t \gets \Phi_{\mathcal{W}_i}^{-1} \left( \mathbf{u}_t^i \right)$, $\mathbf{u}_t^i \in \mathbb{R}^N $

\STATE \textbf{End for}

\STATE \textbf{End}

\STATE \textbf{Outputs}: $\mathbf{W}_t = [\mathbf{w}^1_t \; \mathbf{w}^2_t \; \dots \; \mathbf{w}^M_t] \in \mathbb{R}^{N \times M}$ consisting of samples of random variables with correlation specified by $\mathbf{C}$ ($N$ variables in a column are sampled from the same marginal).
\end{algorithmic}
\end{algorithm}

\subsection{Reliability and risk assessment}

We follow the framework proposed by Stover et al.~\cite{stover2023reliability}. Specifically, risk is defined by a risk triplet: a credible adverse event, the likelihood of the event occurring and the consequence of its occurrence. We consider Level 2 and 3 metrics in this work. Level 2 metrics quantify the probability of failure corresponding to a chosen failure mode (reserve inadequacy, loss of load, etc.). Level 3 metrics consider the consequence cost of the adverse event in addition to the probability of failure. Let $A$ denote a QoI and $E$ denote the reliability threshold, then the probability of adverse event ($A>E$) is expressed as:
\begin{align}
    \mathcal{P}(A, E) = p(A>E),
\end{align}
and can be calculated as:
\begin{align}
    p(A>E) = \frac{1}{N}\sum_{i=1}^N U_i,
\end{align}
where $N$ is the number of MC samples, $U_i=1$ if $A_i>E$, otherwise $U_i = 0$. The consequence (monetary cost) for individual failure event is defined as:
\begin{align}
    R(\Tilde{A}) = \int_0^{\Tilde{A}} C(\Tilde{A}) \; d\Tilde{A}
\end{align}
where $\Tilde{A} = A - E$, and $C(\Tilde{A})$ represents the cost function that depends on $\Tilde{A}$. The associated risk is then calculated as the expectation of consequence of all samples:
\begin{align}
    \mathcal{R}(A, E) = \mathbb{E} \left[ R(\Tilde{A}) \right].
\end{align}
The risk assessment framework above is applicable to various failure modes related to reserve adequacy, supply flexibility, ramp rate, etc. 

\begin{figure}[!ht]
    \centering
    \begin{subfigure}[b]{0.75\linewidth}
        \includegraphics[width=\linewidth]{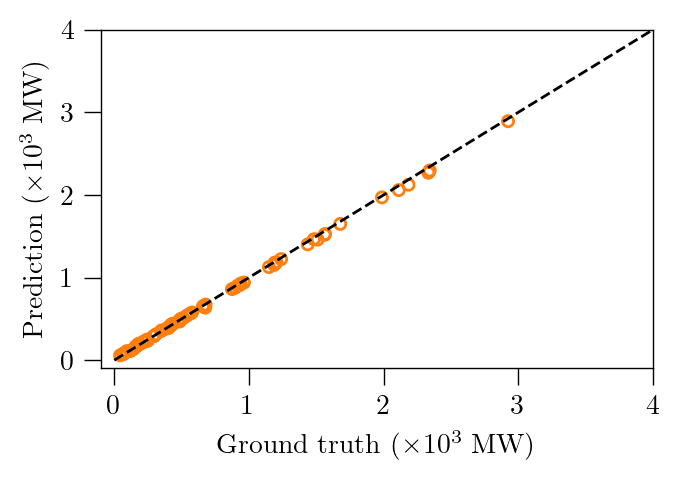}
        \caption{Real-time thermal power generation}
        \label{fig:real_time_PG}
    \end{subfigure}
    \begin{subfigure}[b]{0.75\linewidth}
        \includegraphics[width=\linewidth]{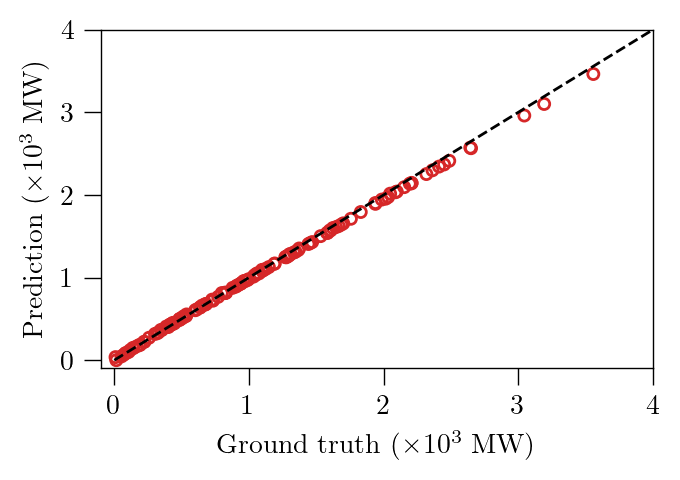}
        \caption{Real-time load shedding}
        \label{fig:real_time_shedding}
    \end{subfigure}

    \begin{subfigure}[b]{0.75\linewidth}
        \includegraphics[width=\linewidth]{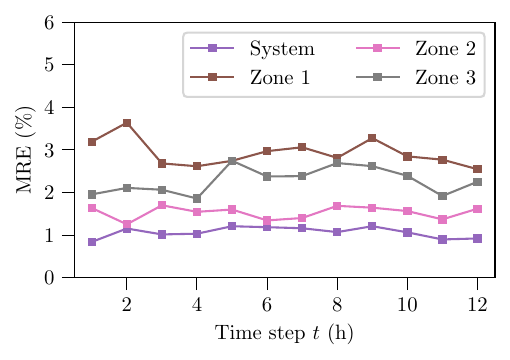}
        \caption{Hours-ahead thermal power generation}
        \label{fig:forward_looking_PG}
    \end{subfigure}
    \begin{subfigure}[b]{0.75\linewidth}
        \includegraphics[width=\linewidth]{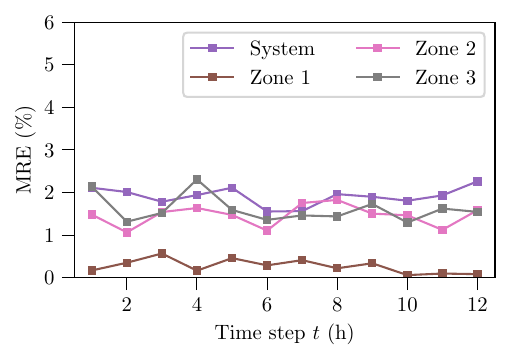}
        \caption{Hours-ahead load shedding}
        \label{fig:forward_looking_shedding}
    \end{subfigure}
    \caption{GNN prediction for real-time and hours-ahead thermal power generation and load shedding. }
    \label{fig:GNN_prediction}
\end{figure}

\begin{figure*}[!ht]
    \centering
    \begin{subfigure}[b]{\linewidth}
        \includegraphics[width=\linewidth]{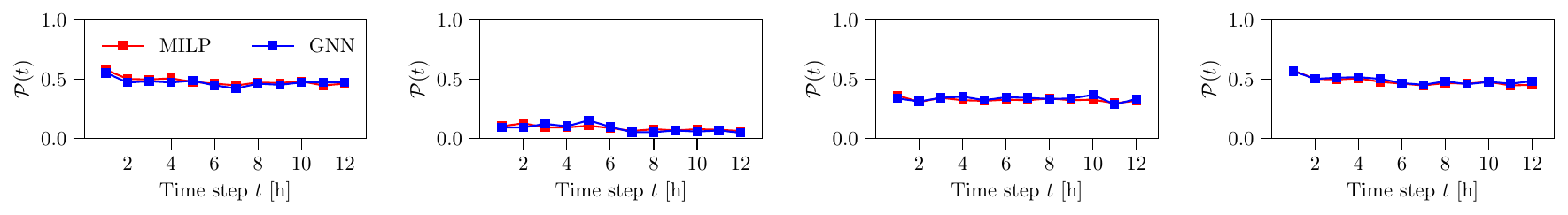}
        \caption{Probability of load shedding. Left to right: system, zone I, II, and III.}
        \label{fig:shedding_probability}
    \end{subfigure}
    \begin{subfigure}[b]{\linewidth}
        \includegraphics[width=\linewidth]{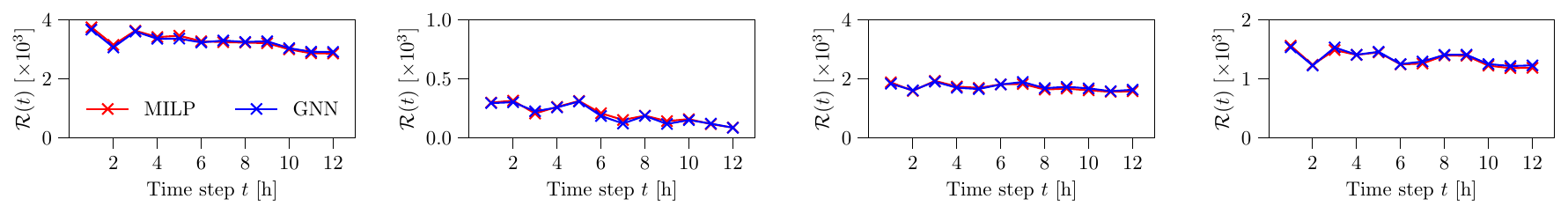}
        \caption{Risk of load shedding. Left to right: system, zone I, II, and III.}
        \label{fig:shedding_risk}
    \end{subfigure}
    \caption{Reliability and risk associated with load shedding at different time steps.}
    \label{fig:forward_looking_reliability_risk}
\end{figure*}

\section{Numerical example}

We use IEEE Case118 power grid (Fig.~\ref{fig:appendix_power_grid}) to perform simulation studies. There are 54 generator buses in this grid, among which 16 are randomly designated as renewable (RES) generators, while the rest are considered as thermal generators. The grid is partitioned into three zones. Wind turbines are used to represent RES generators, wind power will be preferably deployed to meet load demand before thermal plants are taken into consideration. An aggregated load demand and wind power generation are used for each zone. These aggregated values are utilized in modeling the spatio-temporal dependency between different zones, while a distribution sequence is used to determine grid variables at individual bus in the same zone. 

We set $T = 1$ and $T = 12$ hours for real-time (OPF) dispatch and hours-ahead reliability unit commitment (RUC/SCUC) planning, respectively. Stochastic variables are sampled from their marginals. Specifically, wind speed is assumed to follow Weibull distribution (and converted to wind power) while load is described by truncated normal distribution. Parameters for these distributions are summarized in Table~\ref{tab:PDF_parameters}. We use MATPOWER to solve the security-constrained optimization problems by calling state-of-the-art MILP solver Gurobi. Overall, 1000 OPF and SCUC solutions are generated for model training and evaluation.

The GNN surrogates are developed using PyTorch and PyG.  Inputs to the model include load and wind power, grid topology and branch properties, while outputs are thermal power generation and load shedding at system and zone level of the planning period. Data is split into training/validation/testing with proportion at 70\%/10\%/20\%, and the training is conducted on a RTX A6000 graphic card with 24 GB RAM.


\section{Results}

Results are discussed in this section. We first evaluate model performance in terms of thermal power generation and load shedding prediction at system and zonal level. We then examine the accuracy of reliability and risk quantification using GNN predictions.

\subsection{Real-time dispatch}

The comparison between GNN prediction and ground truth of zonal thermal power generation and zonal load shedding is shown in Figs.~\ref{fig:real_time_PG} and \ref{fig:real_time_shedding}. The excellent consistency demonstrates a high degree of accuracy, indicating the GNN model's robust predictive capability. It suggests that the model has successfully captured the structure of power grid and is capable to forecast thermal power generation as well as load shedding.

\subsection{Hours-ahead RUC}

In the hours-ahead RUC, multiple future time steps are of interest. GNN performance is evaluated in terms of mean relative error (MRE) over the entire hours-ahead RUC horizon consisting of 12 time steps. As shown in Fig.~\ref{fig:forward_looking_PG}, the MRE of thermal power generation at each time step within the scheduling horizon remains below 4\%, demonstrating the excellent accuracy. The MRE for thermal power generation at system and zone II is below 2\%. In zone I and III, the MRE is slightly higher but still remains under 3\%. The MRE goes beyond 3\% at only two (out of twelve) time steps. GNN prediction for load shedding is shown in Fig.~\ref{fig:forward_looking_shedding}. Here, the MRE is less than 1\% for zone I and 3\% for other zones as well as the system. Overall, GNN demonstrates excellent predictive capability w.r.t. thermal power generation and load shedding, and thus can be a useful surrogate model for demanding computational tasks in power grid operation like OPF and SCUC.

\subsection{Reliability and risk assessment} 

Here, we present the reliability and risk assessment results based on GNN predictions. For real-time prediction, the results are summarized in Table~\ref{tab:real_time_reliability_risk}. It is observed that the probability of undesirable events (i.e., load shedding) and corresponding risk calculated using GNN predictions show an excellent match with those computed by solving the OPF problem (the ground truth). We notice that load shedding is more likely to happen in zone II ($\mathcal{P} = 0.313$), as compared to zone I and III ($\mathcal{P} =$ 0.203 and 0.256, respectively).
\begin{table}[H]
    \centering
    \caption{Real-time reliability and risk quantification in terms of load shedding at the system and zone levels}
    \begin{tabular}{c|c|c|c|c|c|c|c|c}
        \Xhline{2\arrayrulewidth}
        \multirow{2}{*}{} & \multicolumn{2}{c|}{\textbf{I}} & \multicolumn{2}{c|}{\textbf{II}} & \multicolumn{2}{c|}{\textbf{III}} & \multicolumn{2}{c}{\textbf{System}} \\
        \cline{2-9}
         & \textit{OPF} & \textit{GNN} & \textit{OPF} & \textit{GNN} & \textit{OPF} & \textit{GNN} & \textit{OPF} & \textit{GNN} \\
         \Xhline{2\arrayrulewidth}
         $\mathcal{P}$ & 0.203 & 0.203 & 0.313 & 0.315 & 0.256 & 0.257 & 0.312 & 0.312  \\
         \hline
         $\mathcal{R}$ & 4068 & 4060 & 6264 & 6300 & 5120 & 5130 & 6140 & 6240  \\
         \cline{2-9}
        \Xhline{2\arrayrulewidth}
    \end{tabular}
    \label{tab:real_time_reliability_risk}
\end{table}

With regard to the hours-ahead RUC, the reliability and risk assessment results are summarized in Fig.~\ref{fig:shedding_probability}. GNN predictions match well with ground truth (MILP-based assessment) in both system and zonal level at different times. Specifically, probability of load shedding in zone I is close to zero, noticeably smaller than II and III, which are around 0.4 and 0.5, respectively. The resulting risk quantification is also demonstrated in Fig.~\ref{fig:shedding_risk}. 

\section{Conclusion}

The utilization of GNN as surrogate model for quantification of risk underlying real-time and hours-ahead decision-making in power grid operation is investigated. Large amount spatio-temporally correlated MC samples of stochastic variables (wind power and load demand) are drawn from their joint probability (forecast) distribution and are used to obtain numerical OPF and SCUC solutions. GNN models are trained following supervised learning approach with numerical solutions as training data. Load shedding at system and zonal level is considered as the failure event for reliability and risk assessment. The proposed GNN surrogate-based reliability and risk estimation methodology was demonstrated on the IEEE Case118 power grid. The excellent prediction accuracy and lower computational cost of the GNN model indicate that GNN can be good surrogates for computationally expensive OPF and SCUC problems. Our results also demonstrated that GNN surrogates can provide accurate estimate of both real-time and hours-ahead grid operational risk.

\bibliographystyle{ieeetr}
\bibliography{refs}

\begin{thebibliography}{10}

\bibitem{ghorani2019risk}
R.~Ghorani, F.~Pourahmadi, M.~Moeini-Aghtaie, M.~Fotuhi-Firuzabad, and
  M.~Shahidehpour, ``Risk-based networked-constrained unit commitment
  considering correlated power system uncertainties,'' {\em IEEE Transactions
  on Smart Grid}, vol.~11, no.~2, pp.~1781--1791, 2019.

\bibitem{zheng2019hierarchical}
X.~Zheng, H.~Chen, Y.~Xu, Z.~Liang, and Y.~Chen, ``A hierarchical method for
  robust scuc of multi-area power systems with novel uncertainty sets,'' {\em
  IEEE Transactions on Power Systems}, vol.~35, no.~2, pp.~1364--1375, 2019.

\bibitem{ramesh2021accelerated}
A.~V. Ramesh, X.~Li, and K.~W. Hedman, ``An accelerated-decomposition approach
  for security-constrained unit commitment with corrective network
  reconfiguration,'' {\em IEEE Transactions on Power Systems}, vol.~37, no.~2,
  pp.~887--900, 2021.

\bibitem{kourounis2018toward}
D.~Kourounis, A.~Fuchs, and O.~Schenk, ``Toward the next generation of
  multiperiod optimal power flow solvers,'' {\em IEEE Transactions on Power
  Systems}, vol.~33, no.~4, pp.~4005--4014, 2018.

\bibitem{paturet2020stochastic}
M.~Paturet, U.~Markovic, S.~Delikaraoglou, E.~Vrettos, P.~Aristidou, and
  G.~Hug, ``Stochastic unit commitment in low-inertia grids,'' {\em IEEE
  Transactions on Power Systems}, vol.~35, no.~5, pp.~3448--3458, 2020.

\bibitem{stover2023reliability}
O.~Stover, P.~Karve, and S.~Mahadevan, ``Reliability and risk metrics to assess
  operational adequacy and flexibility of power grids,'' {\em Reliability
  Engineering \& System Safety}, vol.~231, p.~109018, 2023.

\bibitem{zhang2023graph}
Y.~Zhang, P.~M. Karve, and S.~Mahadevan, ``Graph neural networks for power grid
  operational risk assessment,'' {\em arXiv preprint arXiv:2311.03661}, 2023.

\bibitem{wu2020comprehensive}
Z.~Wu, S.~Pan, F.~Chen, G.~Long, C.~Zhang, and S.~Y. Philip, ``A comprehensive
  survey on graph neural networks,'' {\em IEEE transactions on neural networks
  and learning systems}, vol.~32, no.~1, pp.~4--24, 2020.

\bibitem{liao2021review}
W.~Liao, B.~Bak-Jensen, J.~R. Pillai, Y.~Wang, and Y.~Wang, ``A review of graph
  neural networks and their applications in power systems,'' {\em Journal of
  Modern Power Systems and Clean Energy}, vol.~10, no.~2, pp.~345--360, 2021.

\bibitem{kipf2016semi}
T.~N. Kipf and M.~Welling, ``Semi-supervised classification with graph
  convolutional networks,'' {\em arXiv preprint arXiv:1609.02907}, 2016.

\end{thebibliography}

\appendix
\renewcommand\thefigure{\thesection\arabic{figure}}    
\setcounter{figure}{0}   

\renewcommand\thetable{\thesection\arabic{table}}    
\setcounter{table}{0} 
\subsection{Covariance matrix for spatial dependency}
\begin{align}
    C = \begin{bmatrix}
            1 & 0.7 & 0.5 & 0.1 & 0.05 & 0.03 \\
            0.7 & 1 & 0.4 & 0.02 & 0.08 & 0.05 \\
            0.5 & 0.4 & 1 & 0.06 & 0.04 & 0.1 \\
            0.1 & 0.02 & 0.06 & 1 & 0.3 & 0.4 \\
            0.05 & 0.08 & 0.04 & 0.3 & 1 & 0.6 \\
            0.03 & 0.05 & 0.1 & 0.4 & 0.6 & 1 \\    
        \end{bmatrix} \nonumber
\end{align}

\subsection{Power grid used in numerical example}

\begin{figure}[H]
     \centering
     \includegraphics[width=0.8\linewidth]{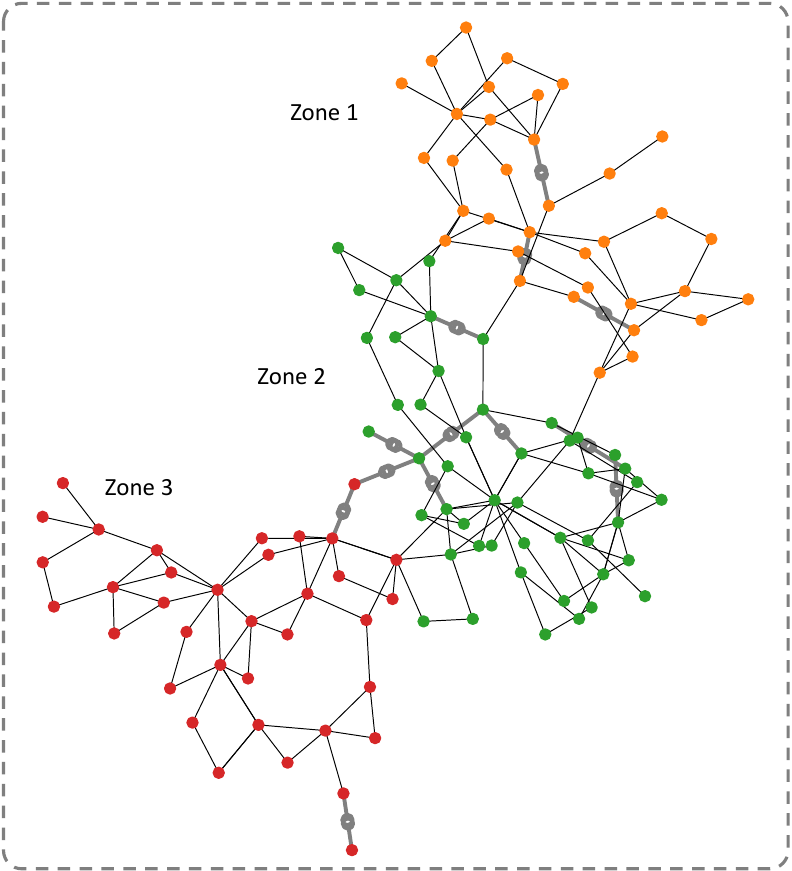}
     \caption{IEEE Case118 power grid.}
     \label{fig:appendix_power_grid}
\end{figure}

\subsection{Parameters of marginal distributions}
\begin{table}[!ht]
    \centering
    \caption{Parameters used for the truncated normal PDF (load, $MW$) and Weibull PDF (wind speed, $m/s$).}
    \label{tab:PDF_parameters}
    \begin{tabular}{c|c|c|c|c|c|c}
    \Xhline{2\arrayrulewidth}
        \multicolumn{2}{c|}{\textbf{Zone/PDF}} & \textbf{Location} & \textbf{Shape} & \textbf{Scale} & \textbf{Left trca.} & \textbf{Right trca.} \\
        \Xhline{2\arrayrulewidth}
        \multirow{2}{*}{I} & TN & 85 & -- & 10 & 55 & 115 \\
        \cline{2-7}
        & WB & 1 & 20 & 11 & -- & -- \\
        \hline
        \multirow{2}{*}{II} & TN & 90 & -- & 12 & 55 & 125 \\
        \cline{2-7}
        & WB & 3 & 15 & 8 & -- & -- \\
        \hline
        \multirow{2}{*}{III} & TN & 95 & -- & 15 & 40 & 150 \\
        \cline{2-7}
        & WB & 1 & 10 & 6 & -- & -- \\
    \Xhline{2\arrayrulewidth}
    \end{tabular}
\end{table}

\end{document}